%% file: main.tex
\documentclass[letterpaper, 10 pt, conference]{ieeeconf}  

\IEEEoverridecommandlockouts                              

\usepackage{amsmath} 
\usepackage{xcolor}
\usepackage{graphicx}
\graphicspath{{./figs/}}
\newcommand{\ie}{i.e.}
\newcommand{\eg}{e.g.}
\input{math_definition}

\usepackage{hyperref}
\usepackage{setspace}
\usepackage{algorithm2e}

\usepackage{caption}
\usepackage{bm}

\title{\LARGE \bf
Sequential Joint Shape and Pose Estimation of Vehicles with Application to Automatic Amodal Segmentation Labeling
}

\author{Josephine Monica$^{1}$, Wei-Lun Chao$^{2}$, and Mark Campbell$^{3}$
\thanks{$^{1},^{3}$Mechanical and Aerospace Engineering Department, Cornell University, NY, USA
        {\tt\small \{jm2684, mc288\}@cornell.edu}
    }
\thanks{$^{2}$Computer Science and Engineering Department, the Ohio State University, OH, USA {\tt\small chao.209@osu.edu}}
}

\begin{document}

\maketitle
\thispagestyle{empty}
\pagestyle{empty}

\input{abstract}
\input{intro}
\input{related}
\input{completion-approach}
\input{amodal-approach}
\input{exp}

\input{disc}

\section*{Acknowledgement}

This research is supported by grants from the National Science Foundation NSF (IIS-1724282, IIS-2107077, OAC-2118240, and OAC-2112606).

{\small
\bibliographystyle{unsrt}
\bibliography{main}
}
\end{document}

%% file: math_definition.tex
\newlength\savewidth

\usepackage{amssymb}
\usepackage{amsmath,amsfonts}
\usepackage{amsopn}
\usepackage{bm} 
\usepackage{multirow}
\usepackage{threeparttable}
\usepackage{flushend}

\newcommand{\ProbOpr}[1]{\mathbb{#1}}

\newcommand{\expect}[2]{%
\ifthenelse{\equal{#2}{}}{\ProbOpr{E}_{#1}}
{\ifthenelse{\equal{#1}{}}{\ProbOpr{E}\left[#2\right]}{\ProbOpr{E}_{#1}\left[#2\right]}}} 
\newcommand{\var}[2]{%
\ifthenelse{\equal{#2}{}}{\ProbOpr{VAR}_{#1}}
{\ifthenelse{\equal{#1}{}}{\ProbOpr{VAR}\left[#2\right]}{\ProbOpr{VAR}_{#1}\left[#2\right]}}} 



%% file: abstract.tex

\begin{abstract}
Shape and pose estimation is a critical perception problem for a self-driving car to fully understand its surrounding environment. One fundamental challenge in solving this problem is the incomplete sensor signal (\eg, LiDAR scans), especially for faraway or occluded objects.
In this paper, we propose a novel algorithm to address this challenge, which explicitly leverages the sensor signal captured over consecutive time: the consecutive signals can provide more information about an object, including different viewpoints and its motion. By encoding the consecutive signals via a recurrent neural network, not only our algorithm improves the shape and pose estimates, but also produces a labeling tool that can benefit other tasks in autonomous driving research. Specifically, building upon our algorithm, we propose a novel pipeline to automatically annotate high-quality labels for amodal segmentation on images, which are hard and laborious to annotate manually. Our code and data will be made publicly available.


\end{abstract}

%% file: intro.tex

\section{Introduction}
A self-driving car must perceive its environment, identify other traffic participants (\eg, vehicles and pedestrians), and importantly, estimate their shapes and poses in order to plan and act safely. 
One of the fundamental challenges for these problems is the sensor signal: a LiDAR scan may only capture one partial view of an object, making shape and pose estimation an ill-posed problem. 
Many existing approaches address this challenge by
training a neural network to encode prior knowledge of complete object shapes~\cite{pcn, pointcompletion-shapeprior,goforth2020joint, weaklyshapecompletion}. While showing promising results, these approaches are still highly sensitive to the quality of input signals.
Specifically, the accuracy of shape and pose estimation drastically drops for faraway or heavily occluded objects whose signals are sparse and limited.

In this paper, we propose to address this challenge by explicitly leveraging consecutive LiDAR scans: we find that existing methods process each LiDAR scan independently, even though the same object may appear consecutively over time.
We argue that consecutive LiDAR scans are crucial for high-quality pose and shape estimation. 
First, while an object may be partially or sparsely observed at each time step, the observations can collectively render a more complete shape over time. Second, traffic participants like vehicles usually move in relatively constrained ways such that temporal information can help correct unreliable pose estimates, as evidenced by the improvement of video-based object detection \cite{yin2020lidar} and object tracking \cite{weng2019baseline,yin2020center} over frame-wise detection. Third, for objects that are rarely seen in the past (\ie, in the training data), the observation over time essentially offers extra data to adapt the algorithm for improved estimation. 

We propose a novel learning-based approach for joint shape and pose estimation that explicitly takes advantage of the consecutive LiDAR scans. Given a sequence of point cloud segments that coarsely captures a single object\footnote{Point cloud segments of individual objects can be extracted from LiDAR scenes through 3D instance segmentation, 3D object detection, or clustering, followed by tracking and data association over time.} (specifically, vehicles in this paper), we propose to fuse the newly extracted features from the current point cloud segment with those from the past via a recurrent neural network~\cite{GRU}. Leveraging past measurements provides additional information, especially for estimating the complete shape.
In addition, the network can internally learn a motion and behavioral model that is beneficial for pose estimation. We demonstrate that our approach in using consecutive LiDAR scans improves the accuracy of shape and pose estimates through validation on both simulated and real datasets.


Besides improving the shape and pose estimation along with other downstream tasks \emph{online}, our sequential approach can also benefit other autonomous driving tasks \emph{offline}.
Specifically, for this paper, we propose to apply the approach to automatically annotate labels for the task of amodal instance segmentation of images, a task aiming to segment the full (amodal) masks of objects irrespective of potential occlusions ~\cite{li2016amodal,zhu2017semantic,xiao2020amodal,zhan2020self}. This is particularly important in autonomous driving, where dense traffic and adverse conditions (\eg, snow, rain, night) may obfuscate objects. 
One bottleneck of this research, aside from the algorithm design, is the lack of large-scale datasets annotated with ground-truth amodal masks.
Concretely, manual annotation of amodal masks is laborious and heavily relies on the subjective intuition of the occluded objects' shape by the annotator, making it hard to maintain consistency and reliability of the labels. 
Fortunately, autonomous driving data is typically collected in sequence, and usually includes both image and LiDAR information. 
Thus, we leverage the point cloud sequence and our sequential algorithm to estimate the complete 3D object shapes at each time frame, these 3D shape results are then projected onto the corresponding image to obtain the amodal masks. The occlusion ordering can be immediately obtained from depth information. Overall, our \textit{automatic} labeling pipeline helps to resolve the ambiguity in manually annotating amodal masks from just a single image.

Our main contributions are:
\begin{itemize}
	\item The first joint shape and pose estimation algorithm that leverages information in time frames to improve sequential estimates.
	\item A simulated dataset of sequences of partial LiDAR measurements of vehicles along with their corresponding complete 3D shapes. The data and data generation code will also be released to enable other researchers to create their own data.
	\item A novel automatic annotation pipeline for amodal instance segmentation that is built upon our sequential algorithm.
\end{itemize}

%% file: related.tex
\section{Related Work}
\subsection{Shape Estimation}
Previous works have proposed different methods and representations in estimating vehicles' shapes from LiDAR observations for fine-grained understanding beyond simple bounding boxes \cite{held2016robust, 7786875, kramerlidar, Monica2020}. However, these works obtain shape estimates only for parts of vehicles that have been previously seen, often resulting in incomplete shape estimates. \cite{kundu20183d, ke2020gsnet} predict the complete shape of an object from an image through a shape  matching process, \ie, representing the object shape as a combination of shape basis, requiring an extensive CAD model library. While 2D image projections of the resulting shape estimates may be accurate, RGB-based methods lack the 3D depth spatial information that is crucial to produce accurate shape estimates in 3D space. As LiDAR sensor is commonly available in autonomous driving setup, 3D point cloud information can be utilized for more accurate shape and pose predictions. 

Point cloud completion aims to predict the complete shape of an object given an incomplete point cloud measurement. Point Completion Network (PCN) \cite{pcn}, one of the pioneers for learning-based shape completion, proposes a simple encoder-decoder shape completion network. \cite{topnet} designs a new decoder architecture that generates point cloud based on a hierarchical tree structure. More recently, \cite{pfnet} proposes a completion method that directly predicts the missing point cloud and appends it to the point cloud input.
 However, they all assume well-aligned, canonical point cloud inputs, which are hard to obtain for actual sensor data. Thus, these methods require a pose estimation and a point cloud alignment to a canonical pose as preprocessing steps to operate on real-world data. Consequently, the shape estimation is subject to errors from the pose estimation. 
To address this issue, \cite{goforth2020joint} proposes a network that jointly estimates the pose and complete shape of vehicles, sharing information between the two tasks. Our work builds upon these efforts. However, unlike existing works which make an independent estimate for each point cloud input, our method fuses information over time to produce more accurate shape and pose estimates.

\subsection{Pose Estimation}
Pose estimation in autonomous driving refers to inferring the location and orientation of an object. Many detection and tracking algorithms represent vehicles by bounding boxes (parameterized by sizes, orientations, and center locations) and perform traditional filtering, such as the Kalman filter, to fuse new measurements with prior beliefs \cite{KF-tracking}. However, representing vehicles by bounding boxes eliminates detailed shape information that can facilitate accurate and robust tracking. 
To make full use of shape information, some trackers \cite{ICP-tracking} apply Iterative Closest Point (ICP) to align point cloud measurements. However, they heavily rely on good initialization and often get stuck in local minima. As opposed to ICP, Annealed Dynamic Histogram (ADH) tracker \cite{held2016robust} globally explores the state space to find the best Markovian alignment between measurements. However, ADH scores the point cloud alignment only based on the latest two measurements. Thus, ADH lacks robustness when the point cloud is sparse, or the viewpoint changes rapidly. Learning-based approaches \cite{AlignNet3D,itn,deepgmr,pointvotenet} benefit from curated training data and can learn to be robust against occlusions and inferior sensor data. However, most of these approaches operate separately on individual frames and do not utilize past information along tracks.

\subsection{Amodal Mask Labeling}
Labeling amodal masks is commonly deemed as a subjective and ill-posed problem, as it involves the annotators' intuition to predict the occluded masks. Thus, it requires rigorous guidelines and control to ensure the consistency and quality of the labels. 
For example, the KINS dataset \cite{qi2019amodal} provides manually annotated amodal mask labels on the autonomous driving KITTI object detection data \cite{geiger2013vision}. To obtain high-quality labels, the data is repetitively labeled six times by three different annotators. This shows the difficulty of amodal mask labeling.
 
Several attempts have been made to automate the amodal mask label generation. \cite{li2016amodal} intentionally introduces occlusion to the image by overlaying unoccluded instances with other object masks. The new overlaid image is used as the input data, while the original unoccluded masks serve as the amodal mask labels. While this can efficiently provide accurate amodal mask labels, the generated images may not look realistic or reflect a realistic occlusion relationship in the real world. \cite{kar2015amodal} aligns 3D CAD models from PASCAL 3D+ dataset \cite{xiang2014beyond} to the target instances in PASCAL VOC \cite{everingham2011pascal} images and projects the model onto the image frames to acquire the labels. While the generated images come from a real-world occlusion relationship, the process requires object 3D CAD models and manual alignment of the models.

%% file: completion-approach.tex
\begin{figure*}[t]
	\centering
	\includegraphics[width=1.9 \columnwidth]{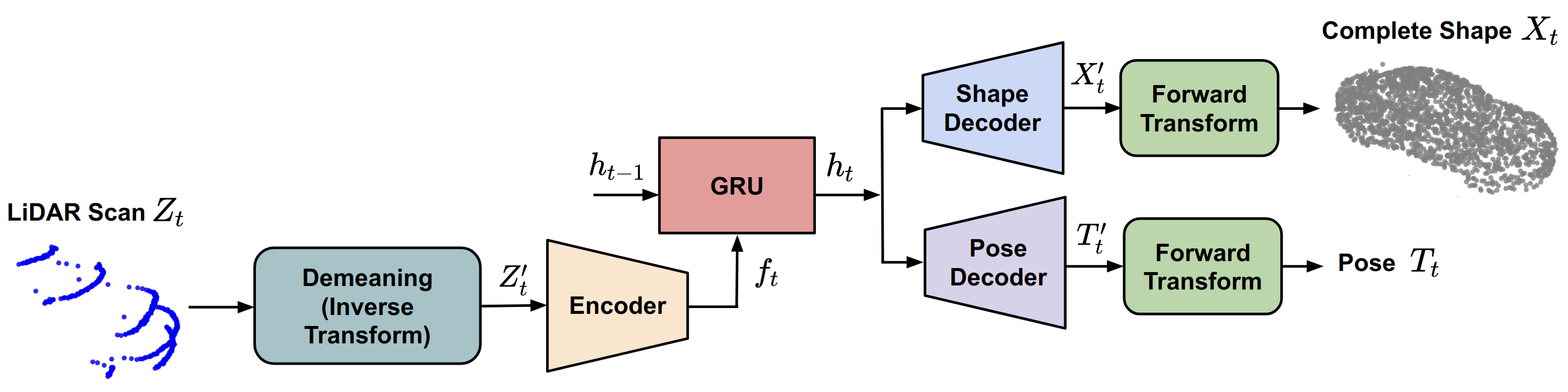}
	\caption{Overview diagram of our sequential shape and pose estimation pipeline. The encoder consists of two stacked PointNet \cite{pointnet} layers. The fusion network is a single layer GRU. The shape and pose decoders are Multi-layer Perceptrons (MLPs). \label{fig:overview}}
\end{figure*}
\section{Sequential Joint Shape and Pose Estimation}
We formulate the problem of \emph{sequential joint shape and pose estimation} as estimating the complete point cloud $X_t$ and homogeneous transformation $T_t$ (from the canonical reference pose) of an object at time $t$, given unaligned partial point cloud measurements $Z_{1:t}$ of the same object. As vehicles commonly move on a planar ground, we only estimate the planar translation and rotation for $T_t$. $Z_{1:t}$ can be obtained by a 
frame-wise point cloud segmentation (\eg, via object detection \cite{shi2019pointrcnn,qi2018frustum,lang2019pointpillars,zhou2018voxelnet} or 3D instance segmentation \cite{wang2019ldls,wu2018squeezeseg}), followed by a standard data association over time \cite{tracking-association,weng2019baseline,yin2020center}. Our approach outputs the complete point cloud in the original pose of the unaligned measurement $Z_t$ rather than in its canonical pose.

\subsection{Pipeline}
Our algorithm begins with estimating the complete shape and pose of a vehicle from its first observation using an encoder-decoder-style neural network inspired by \cite{goforth2020joint}. For the sequential time frames, however, unlike existing work, we recursively fuse the newly extracted point cloud features with those from the past through Gated Recurrent Units (GRUs) \cite{GRU}.
\autoref{fig:overview} shows the overall pipeline of our sequential shape and pose estimation framework. First, we preprocess the input by shifting the partial point cloud measurement $Z_t$ by its mean $\bar{Z}_t$ (demeaning in \autoref{fig:overview}):
\begin{equation}
Z_t' = Z_t -\bar{Z}_t
\end{equation}
In our pipeline, demeaning assists the fusion process by coarsely aligning point cloud measurements from different time steps to the same origin coordinate. Additionally, this step makes training and inference easier by narrowing the network input and output range around the origin. The demeaned point cloud $Z_t'$ is then passed to the joint encoder to extract the feature vector $f_t$:
\begin{equation}
f_t = f(Z_t')
\end{equation}
which summarizes the measurement at time $t$.

Instead of using $f_t$ alone to predict the complete shape and pose, our method leverages all measurements $Z_{1:t}$ up to the latest frame in making the prediction. In particular, we employ a GRU module which recursively updates its hidden state $h_t$ by fusing the previous hidden state $h_{t-1}$ with the current measurement feature $f_{t}$:
\begin{equation}
h_t = GRU(h_{t-1}, f_t)
\end{equation}
The hidden state $h_t$, which can be seen as a summary of all available measurements $Z_{1:t}$, is then used as the input to the pose and shape decoders.
This fusion process is executed at the feature level instead of the point-cloud level, allowing the network to learn better how to combine and extract the most useful information from $Z_{1:t}$.

The updated hidden state $h_t$ is then passed to the shape and pose decoders to estimate the complete point cloud $X_t'$ and pose $T_t'$: 
\begin{equation}
X_t' = g_{\text{shape}}(h_t), \; T_t' = g_{\text{pose}}(h_t)
\end{equation} These shape and pose estimates are still in the $Z'_t$ coordinate frame (\ie, coordinate frame whose origin lies at the measurement mean). A forward transformation is applied to bring each back to the original measurement $Z_t$ frame:
\begin{equation}
X_t = X_t' + \bar{Z}_t, \; T_t = \begin{bmatrix}
I & \bar{Z}_t \\ 0 & 1
\end{bmatrix}T_t'
\end{equation}

Overall, our pipeline shares some conceptual similarities to shape estimation of a moving object using traditional filtering methods \cite{7786875, Monica2020, wyffels2015joint} which consists of dynamics and measurement update steps. Dynamics update aims to transform the shape estimate to the same coordinate frame as the new measurement to assist the measurement update (\eg, point cloud fusion process). Our demeaning block is analogous to the dynamics update, but instead of transforming the shape estimate forward, it transforms the new measurement back to the previously fused state frame. We design the dynamics update in reverse to avoid having to decode $h_{t-1}$ to shape and pose before transforming then encoding them again to get the next time step hidden state $h_{t}$. This redundant decoding-encoding step will induce information loss and produce sub-optimal results. Meanwhile, our GRU fusion is analogous to the measurement update, which utilizes the measurement information to update the shape and pose estimates. 

\subsection{Training}
Neural networks that contain multiple components are challenging to train in a naive end-to-end manner.
We thus follow \cite{goforth2020joint} to employ a multi-stage training procedure.
\subsubsection*{Stage 1} We train the encoder, GRU, and shape decoder by minimizing the Chamfer Distance (CD) shape loss \cite{shapeloss} between the point cloud estimate $X$ and ground truth $X^\text{gt}$:
\begin{align}\label{eqn:CDLoss}
\begin{split}
L_{\text{CD}}(X,X^{\text{gt}}) = \frac{1}{|X|} \sum_{\hat{x} \in X} \min_{x \in X^{\text{gt}}} \| \hat{x}-x \|_2 \\
+ \frac{1}{|X^{\text{gt}}|} \sum_{x \in X^{\text{gt}}} \min_{\hat{x} \in X} \| x-\hat{x} \|_2 
\end{split}
\end{align}
\subsubsection*{Stage 2} As features from previous shape training have captured some notions about pose \cite{goforth2020joint}, we train the pose decoder on top of those features while freezing the rest of the network weights. We train the pose decoder by minimizing the pose loss \cite{poseloss}: 
\begin{equation}\label{eqn:poseloss}
L_{P}(T,T^{\text{gt}};X^{\text{gt}}) = \frac{1}{|X^{\text{gt}}|} \sum_{x \in X^{\text{gt}}} \lVert T^{-1} x - ({T^{\text{gt}}})^{-1}x \rVert_2^2
\end{equation}
where $T$ and $T^{\text{gt}}$ are the estimated and ground truth transformations.
This loss function penalizes both the translation and rotation errors.
\subsubsection*{Stage 3} Finally, we jointly train the entire network for sequential shape and pose estimation using the following joint loss $L_\text{J}$ \cite{uncertatinty-weight}:
\begin{equation}
L_{\text{J}} = \frac{1}{2 \sigma_{\text{CD}}^2} L_{\text{CD}} + \frac{1}{2 \sigma_{\text{P}}^2} L_\text{P} + \log({\sigma_{\text{CD}}\sigma_{\text{P}}})
\end{equation}
where $\sigma_{\text{CD}}$ and $\sigma_{\text{P}}$ are learnable parameters, denoting the uncertainty of each prediction task. By training the two tasks simultaneously, learning in one
task benefits the other, since the tasks are coupled. Moreover, the network can learn a more general representation capturing both tasks and is less likely to overfit.

%% file: amodal-approach.tex
\section{Automatic Amodal Segmentation Labeling}\label{sec:amodal-approach}
We extend our sequential shape and pose estimation algorithm in point clouds
to automatic amodal segmentation labeling in images, building on the following key observation.
An object instance commonly appears in multiple frames, as driving data is collected in a continuous sequence. Thus, a part of an amodal mask occluded in one frame may be visible in other frames. If we can fuse the information from multiple frames, we can produce amodal masks that are consistent across time frames. 
One naive way is to fuse 2D inmodal masks in the image space --- 2D inmodal masks are easier to annotate or can be obtained from existing segmentation algorithms like Mask-RCNN~\cite{he2017mask}. This method, however, has several challenges. First, an object's 2D size and shape vary across frames depending on its relative pose to the camera. Second, the mask shape is also influenced by the object's 3D orientation, a piece of information that is missing in 2D. Third, while multiple frames provide extra information about an object, some parts may never be seen in any frames.

To solve these issues, we propose to fuse the information over time in 3D by taking advantage of 1) the 3D LiDAR signal that is commonly available in the autonomous driving configuration, and 2) our learned network for shape completion that encodes the prior of complete shapes from training data. 
We feed the point cloud sequence into our sequential point completion network in order to obtain the full amodal information of the object. The complete point cloud is then projected to the image space and post-processed to get an alpha shape as the amodal mask label. Finally, the occlusion ordering is reasoned from depth information.

We present two different scenarios where our automatic labeling pipeline can be used. The first assumes the availability of 3D bounding box ground truth and association, which are typically available on a public driving dataset, on the data to be labeled. In this case, the 3D bounding boxes are used to segment the point cloud objects to get a sequence of partial point cloud objects containing only the target vehicle. Given a segmented point cloud sequence and ground truth transformations, one can simply accumulate these segmented point clouds using the ground truth transformations and use the symmetrical property of vehicles to mirror the point cloud with respect to its heading axis in order to obtain the amodal shape information. However, this may still result in incomplete amodal information for many cases, as will be shown in \autoref{exp-amodal}. So instead, we leverage our sequential point completion network to fuse and complete the amodal information. As our goal is labeling, we can train and evaluate the network on the same dataset.

In the second scenario, we handle situations where 3D bounding box labels are unavailable. In this case, we deploy a pretrained Mask-RCNN \cite{he2017mask} inmodal instance segmentation and LDLS \cite{wang2019ldls} (a 2D to 3D label diffusion framework) to segment the point cloud, followed by data association. Since 3D bounding box labels are not available, we deploy our sequential completion network that is pretrained on another dataset, \eg, synthetic dataset.

%% file: exp.tex
\section{Experiments}
We first evaluate the quality of the completed point cloud and pose estimates on the synthetic and Argoverse \cite{argoverse} real dataset. The translation and rotation errors are computed to evaluate the pose estimate quality. Besides Chamfer Distance (CD), Earth Mover Distance (EMD) \cite{poseloss, MSN-PCN} is also computed to evaluate the completed point cloud through different aspects. The CD captures the global structure of the point cloud, while the EMD captures the point density.

Next, we evaluate our automatic amodal labeling pipeline on the KITTI tracking dataset, comparing our generated masks to the manually annotated labels from KINS \cite{qi2019amodal} by computing the mean intersection over union (mIoU), as well as \% miss, \ie, the percentage of instances that do not have a matching or IoU $<$ 0.5.

\subsection{Data Generation}
 The synthetic dataset can generate perfect and complete point clouds required for training and evaluation. However, the synthetic dataset does not contain real-world challenges such as occlusions, contamination from ground points, LiDAR time synchronization issues, and data mislabeling. Thus, we complement our analysis by training and evaluating on the real-world Argoverse dataset \cite{argoverse}. 

\subsubsection{Synthetic Dataset}
We obtain 183 vehicle CAD models from \cite{CAD-toronto} and randomly split them to 168/15 for training and validation. Each CAD model is fit into 11 different trajectories, where each may contain a different number of frames, resulting in 1848/165 tracks for training and validation. To mimic the real-world behavior of vehicles, we use the trajectory log from the Argoverse dataset \cite{argoverse}.
To simulate partial point cloud measurements, we transform the CAD models to the desired poses in the track and apply raytracing using the Trimesh library \cite{trimesh}. We generate a synthetic dataset with VLP-16 LiDAR configuration placed at a 2m height from the ground. To generate a complete point cloud shape, we use Open3D library \cite{open3d} to sample points from the CAD models and remove interior points that are not visible from the outside.

\subsubsection{Argoverse}
We evaluate our results on car objects in the Argoverse tracking dataset \cite{argoverse} split into 2298/638 tracks for training and validation, where each track contains an average of 70 frames. As we do not focus on segmentation and data association problems, ground truth bounding boxes and track ids are used to crop the LiDAR scene.

While ground truth pose is provided by Argoverse, complete point clouds are not available. Thus, to acquire a shape reference, LiDAR point clouds of vehicles are accumulated over multiple frames. Furthermore, since vehicles are usually symmetrical about their heading axes, we mirror the points along the symmetrical heading axes \cite{lidarsim}. 
Note that even after the LiDAR point clouds are accumulated and mirrored, the developed shape reference may still be non-ideal (\eg, displaying noise, inaccuracy, incompleteness, and non-uniform point density), as it is built from measurements.
As EMD is highly sensitive to these properties, evaluating EMD against such point cloud reference is not meaningful. Thus, we do not report the EMD for Argoverse evaluation.
\subsection{Shape and Pose Estimation}
We run our method recursively from each vehicle's initial to final detection. The shape and pose errors are evaluated at each time step. 
We compare our method to the joint shape and pose estimation from \cite{goforth2020joint} as a baseline. Different from our proposed approach, the baseline makes a prediction on each frame independently.
\subsubsection{Results on Synthetic Dataset}
\autoref{tab:synthetic} shows the
sequential shape and pose estimation results on the synthetic dataset. Our method outperforms the baseline in both shape completion and pose estimation tasks. While CD is the direct metric for training, EMD quantifies the quality of point density. The reported EMD value is much higher than CD due to requiring one-to-one correspondences between points. Nevertheless, our method still outperforms the baseline even on the EMD metric.

\begin{table}[b] 
	\caption{Quantitative result on the synthetic dataset.}
	\label{tab:synthetic}
	\begin{center}
		\begin{tabular}{|c|c|c|c|c|}
			\hline
			\multirow{2}{*}{Method} & \multicolumn{2}{|c|}{Shape Error} & \multicolumn{2}{|c|}{Pose Error} \\ \cline{2-5}
            & CD & EMD & Translation & Rotation\\ \hline
            Baseline& 3.1 cm & 0.50 m& 12.2 cm& 20.2$^o$\\
			\hline
			Ours & 2.3 cm& 0.25 m& 9.4 cm& 12.0$^o$\\
			\hline
		\end{tabular}
	\end{center}
\vskip -5pt
\end{table}

\begin{figure}[t!] 
	\centering
	\includegraphics[width=0.9 \columnwidth]{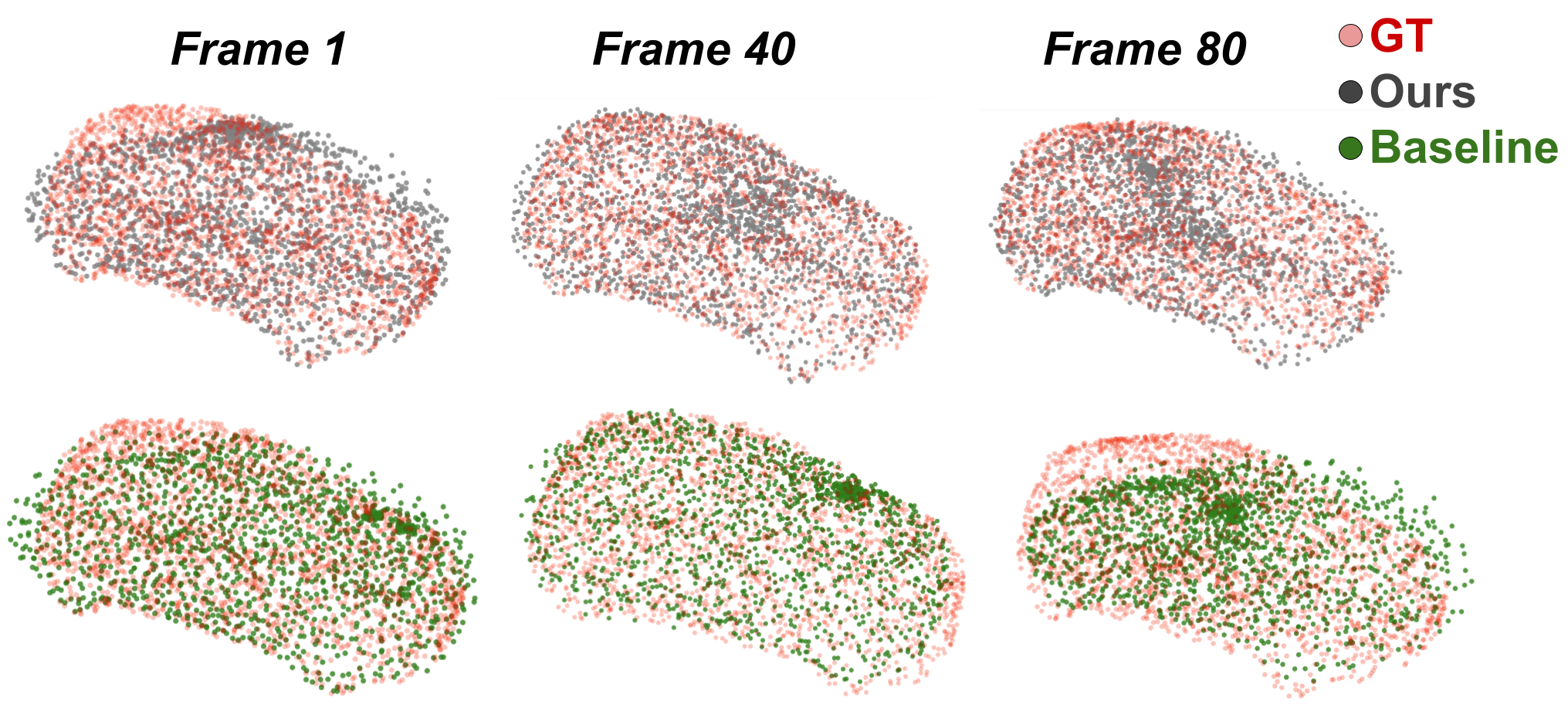}
	\caption{Qualitative shape completion results on a track from the synthetic data over time. Top: Ours. Bottom: Baseline. \label{fig:synthetic-qualitative}}
\vskip -5pt
\end{figure}
\begin{figure}[b!] 
	\centering
	\includegraphics[width=.87 \columnwidth]{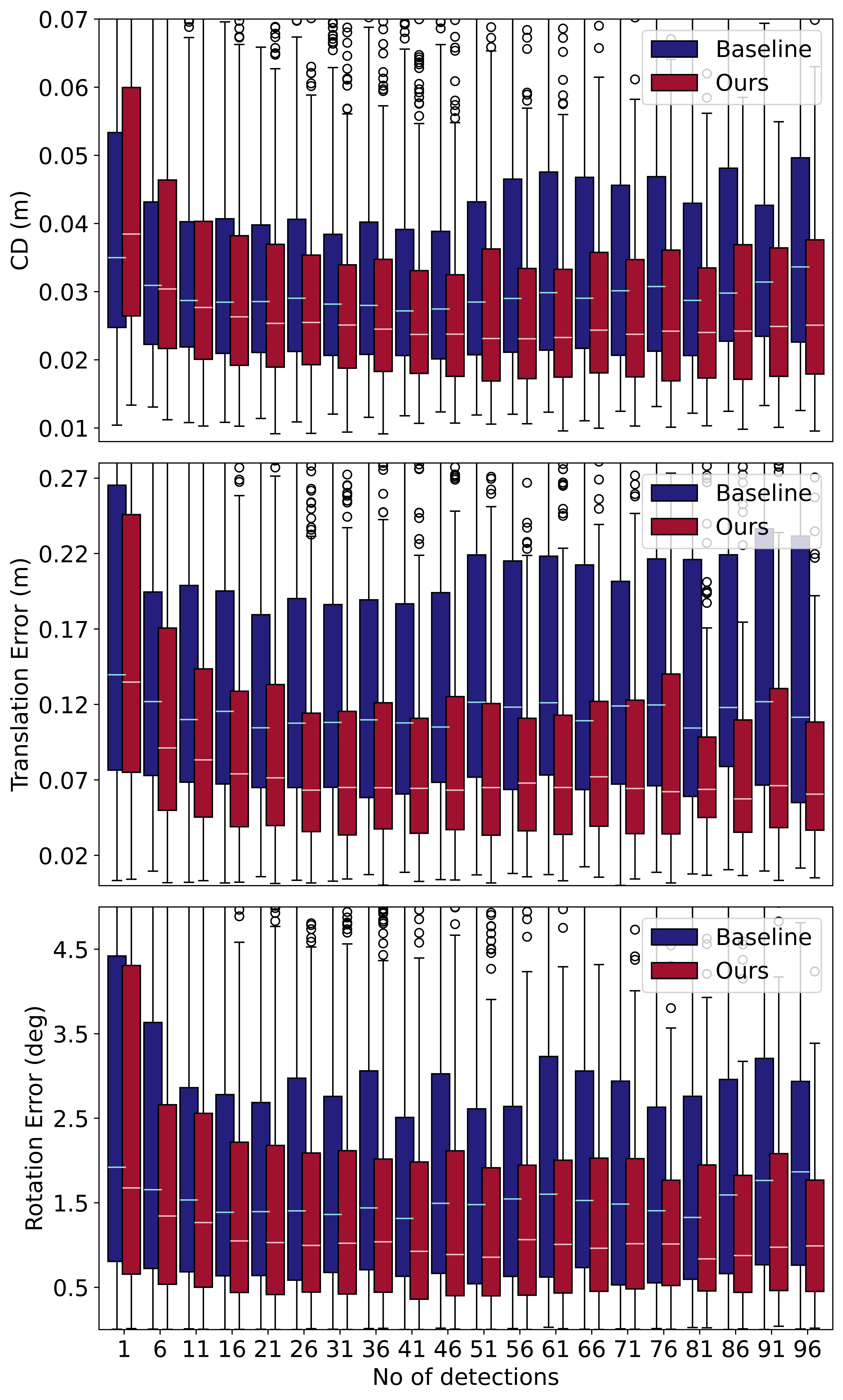} 
	\caption{Statistical error analysis on the shape and pose accuracy as a function of number of detections on the Argoverse dataset. Lower is better. \label{fig:argo_result}}
	\vskip -5pt
\end{figure}
\begin{figure*}[t!] 
	\centering
	\includegraphics[width=1.92 \columnwidth]{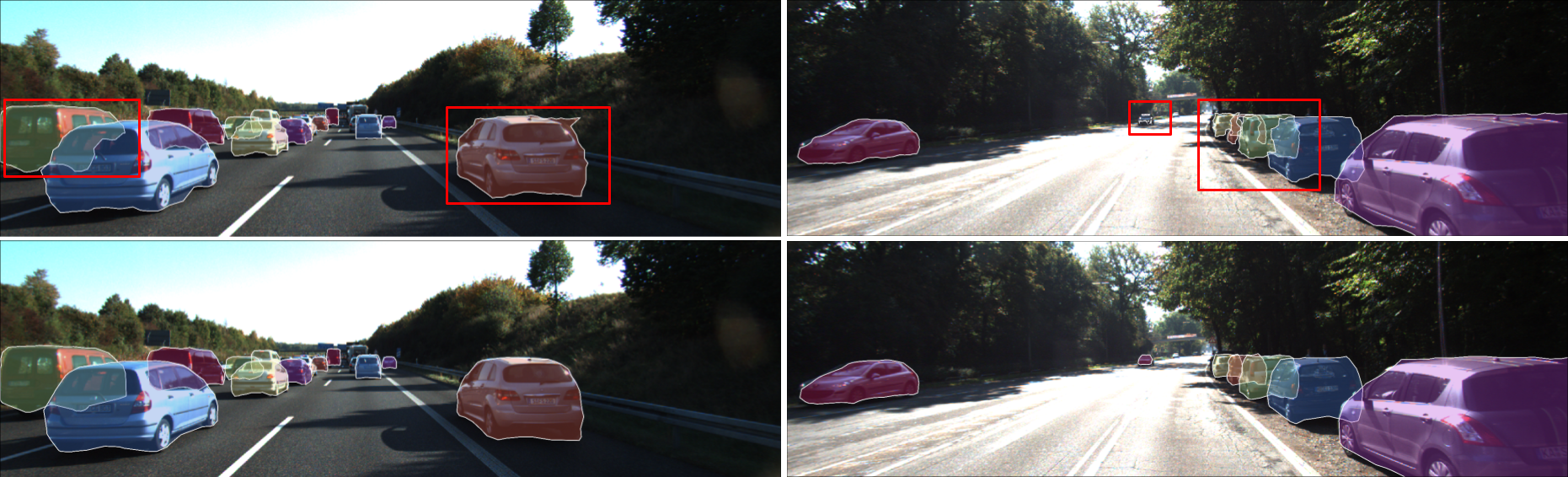}
	\caption{Amodal mask labels from our proposed pipelines. Top: GT-Accumulation. Bottom: GT-SC. The masks inside the red boxes highlight the benefit of using our sequential completion network. \label{fig:amodal-comparison}}
\vskip -5pt
\end{figure*}

To complement the quantitative analysis, shape estimate results at three different time steps are shown in \autoref{fig:synthetic-qualitative}. In this case, the sensor data is initially (frame 1) very sparse. As a result, both methods do not produce accurate shape estimates; they also mispredict the car's orientation by roughly 180$^o$. At frame 40, the sensor data is denser, and both methods produce more accurate shape estimates. Nevertheless, our shape estimate still fits the ground truth better. At frame 80, the sensor data becomes sparse again. As a result, the baseline approach produces an inaccurate shape estimate and a significant orientation error, as seen from \autoref{fig:synthetic-qualitative}. Yet, at this time, our method produces an accurate shape estimate despite the challenging sensor measurement. This shows that by leveraging temporal information, we can produce more accurate and consistent estimates over time.

\subsubsection{Results on Argoverse Dataset}
\autoref{fig:argo_result} plots the shape and pose estimate error metrics with respect to the number of detections. Our method outperforms the baseline in all metrics and shows a general trend of improvement in the quality of shape estimates with more detections. This aligns with our initial claim that performance improves with more frames if fused properly.
\subsection{Automatic Labeling of Amodal Segmentation on KITTI}\label{exp-amodal}
We compare our automatically-generated label qualities under different ablations and cases to KINS manual annotation \cite{qi2019amodal} in \autoref{tab:amodal-mask}.
KINS-manual mIoU represents the consistency of manually annotated labels across different annotators, denoting the quality of the manual labels. 
 GT-Accumulation is our method of accumulating and mirroring partial point cloud sequences. SC-GT refers to our improvement of GT-Accumulation by utilizing our sequential completion. Finally, SC-MRCNN refers to sequential completion using Mask-RCNN~\cite{he2017mask} and label diffusion when 3D ground truth bounding boxes are unavailable.
 
In all three cases (with and without 3D bounding box labels), our proposed labeling approaches achieve comparable mIoU consistency to human-level performance, i.e., KINS cross-annotators consistency. In the case where 3D ground truth bounding boxes are available, running through sequential completion (SC-GT) yields less missing instance match than just accumulating and mirroring (GT-Accumulation). It is important to note that we compare our generated labels to the manually annotated KINS labels and not to the actual amodal mask ground truth that is practically unavailable. The KINS label itself has some noise, \ie, its internal consistency across different annotators is \nolinkurl{~}0.8. Thus, achieving a higher mIoU than that does not necessarily imply a better quality mask.
\begin{table}[htbp] 
	\caption{Amodal mask consistency.}
	\label{tab:amodal-mask}
	\begin{center}
		\begin{tabular}{|c|c|c|c|c|}
			\hline
			Method & 3D GT label& \% Miss & mIoU  \\
			\hline
			KINS-manual& - & - & 0.809\\
			\hline
			GT-Accumulation & Available & 2.69 & 0.813 \\
			\hline
			SC-GT & Available & 0.90 & 0.813 \\
			\hline 
			SC-MRCNN & No & 5.53 & 0.788 \\
			\hline
		\end{tabular}
	\end{center}
\vskip -5pt
\end{table}

To give a better insight into the advantage of our sequential completion network, we qualitatively show several amodal mask examples in \autoref{fig:amodal-comparison}. Specifically, note the area inside the red boxes. We observe cases where merely accumulating and mirroring (GT-Accumulation) is not enough to cover the complete amodal information. Additionally, GT-Accumulation is prone to error or noise in the point cloud, as shown in the white car inside the red box in \autoref{fig:amodal-comparison} top left. On the other hand, despite the noise, our sequential completion network has a smoothing effect that filters out this noise.

Finally, in the absence of 3D ground truth bounding boxes, we can not perform our accumulation and mirroring method (GT-Accumulation). But our sequential completion network (SC-MRCNN) still provides comparable quality masks to human-annotated labels despite the absence of any labels. Although as expected, the performance of SC-MRCNN is slightly worse than SC-GT. This is because segmenting out point cloud using Mask-RCNN and LDLS is not as accurate as using human-annotated 3D ground truth bounding boxes. Additionally, due to the absence of 3D ground truth bounding boxes, the sequential completion network in SC-MRCNN is trained on the synthetic data, which introduces some domain adaptation gap when applied to the KITTI data.

%% file: disc.tex
\section{Conclusion}
We propose a learning-based approach for joint point completion and pose estimation of vehicles from LiDAR point cloud measurements. Uniquely, our method explicitly leverages the temporal information of tracks. We evaluate our method on synthetic and real-world datasets, showing better performance in both shape and pose estimation tasks against the baseline approach. We demonstrate that properly fusing extra temporal information benefits shape and pose estimation. Finally, we propose a novel automatic amodal labeling pipeline using our sequential completion network. 
We evaluate our automatic amodal labeling pipeline and demonstrate comparable quality to human annotations.